\documentclass[11pt,letterpaper]{article}
\usepackage{naaclhlt2016}
\usepackage{times}
\usepackage{latexsym}
\usepackage{hyperref}
\usepackage{csquotes}
\usepackage{graphicx}
\usepackage{pifont}
\usepackage{multirow}
\usepackage{xcolor}
\newcommand{\cmark}{\ding{51}}%
\newcommand{\xmark}{\ding{55}}%

\hypersetup{
    colorlinks=true,
    linkcolor=blue,
    urlcolor=cyan,
    citecolor=blue
    }
    
\naaclfinalcopy % Uncomment this line for the final submission
\title{Are LLMs the Master of All Trades? : Exploring Domain-Agnostic Reasoning Skills of LLMs}

\author{Shrivats Agrawal\\
	    University of Pennsylvania\\
	    {\tt shriv9@seas.upenn.edu}
	  }

\date{27 Feb, 2023}
\begin{document}
\maketitle
\begin{abstract}
The potential of large language models (LLMs) to reason like humans has been a highly contested topic in Machine Learning communities. However, the reasoning abilities of humans are multifaceted and can be seen in various forms, including analogical, spatial and moral reasoning, among others. This fact raises the question whether LLMs can perform equally well across all these different domains. This research work aims to investigate the performance of LLMs on different reasoning tasks by conducting experiments that directly use or draw inspirations from existing datasets on analogical and spatial reasoning. Additionally, to evaluate the ability of LLMs to reason like human, their performance is evaluted on more open-ended, natural language questions. My findings indicate that LLMs excel at analogical and moral reasoning, yet struggle to perform as proficiently on spatial reasoning tasks. I believe these experiments are crucial for informing the future development of LLMs, particularly in contexts that require diverse reasoning proficiencies. By shedding light on the reasoning abilities of LLMs, this study aims to push forward our understanding of how they can better emulate the cognitive abilities of humans. The code developed for conducting these experiments can be found \href{https://github.com/ShrivatsAgrawal/Are-LLMs-the-Master-of-All-Trades-Exploring-Domain-Agnostic-Reasoning-Skills-of-LLMs}{here}.

\end{abstract}

\section{Introduction}
With the emergence of Large Language Models (LLMs) we have been able to achieve unprecedented performance in Natural Language Processing tasks and have also enabled machines to generate human-like text and perform complex language-based tasks \cite{bert,gpt2,gpt3}. 
However, with the rising difficulty in distinguishing content generated by LLMs from that written by humans, the central question of whether LLMs can truly emulate human reasoning and perform reasoning across a broad range of domains is now more relevant than ever. \cite{reason1}

At the heart of every human cognitive activity like problem solving and decision-making lies reasoning of some form. Yet, human reasoning is multifaceted and can be seen in diverse forms such as analogical, spatial, causal, and moral reasoning, among others \cite{reason2}. Therefore, studying the extent to which LLMs can perform equally well across these different domains of reasoning is essential to advancing our understanding of these models and their potential applications in real world settings.

Humans have the ability to reason and understand things by using analogies, even when presented with textual information alone \cite{analogical1}. However, the development of spatial reasoning skills requires an approach driven by input from other senses, with humans needing to interact with their environment using senses like vision, sound, and touch. Meanwhile, moral reasoning abilities are primarily shaped by a person's education, experiences, and surroundings. The clear differences between these types of reasoning motivated me to focus my attention and assess the reasoning abilities of LLMs on analogical, spatial, and moral reasoning tasks.
 My study employs a range of experimental methods, including the use of existing datasets directly, such as  BATS \cite{bats}, as well as constructing a toy dataset by drawing inspiration from SpartQA \cite{spartqa} to directly evaluate LLMs abilities in analogical and spatial reasoning. I also evaluate LLMs performance on more open-ended, natural language questions that probe into these reasoning domains, as well as moral reasoning. My findings reveal that LLMs exhibit impressive capabilities in analogical and moral reasoning, yet face challenges in performing as proficiently on spatial reasoning tasks.

% The structure of the remaining paper is as follows. In Section 2, the experimental methodology for each specific type of reasoning along with the relevant datasets is outlined. Thereafter, experimental results, including both success and failure cases, followed by relevant analysis is presented in Section 3. Finally, the paper is concluded in Section 4, highlighting the significance of this research in advancing our current understanding of LLMs reasoning capabilities in emulating human intelligence. It is my sincere hope that by combining both dataset-centric and free-form natural language based questions approach, this work will help inform the development of more sophisticated and \textit{intelligent} artificial reasoning systems.

\section{Experimental Setup}
The experimental methodology for this paper is partitioned into three distinct sections, each pertaining to one of the three reasoning domains evaluated: analogical reasoning, spatial reasoning, and moral reasoning.
\subsection{Analogical Reasoning}
To evaluate LLMs' analogical reasoning capabilities, a subsection of the BATS dataset \cite{bats} is used along with free-form natural language questions gathered from the web and some created in-house. GPT-3 \textit{davinci-003} \cite{gpt3} is used for performing experiments with the controlled dataset and ChatGPT is used for conversational style natural language questions.
\subsubsection{Controlled Dataset}
An experimental dataset is created for analogical reasoning by randomly sampling from the lexicographic semantics portion of the BATS dataset. This section of the dataset is chosen as previous models have shown the least accuracy on this section and it is the hardest to solve as the analogies don't follow any explicit grammatical rules. To ensure a uniform evaluation across the entire lexicographic section, 20 random analogies are sampled from each of the 10 subfiles of lexicographic semantics, resulting in a total of 200 analogies. The task is framed as a cloze question test, where the model is required to fill in the blank. For instance, the model is prompted as follows for one such example: \\
``Fill in the blank: \underline{\textit{Sofa}} is to \underline{\textit{piece of furniture}} as \underline{\textit{stapler}} is to \underline{$\;\;\;\;\;\;\;$}.$\;\;$Answer:$\;$''\\
The models are evaluated by considering exact matches in the set of provided answers as correct and incorrect otherwise.

\subsubsection{Conversational Prompts}
To evaluate LLMs in a way similar to humans, I collated a dataset containing free-form natural language questions. The dataset consists of questions like \textit{``Explain how a camera works by taking the example of an eye.''} and \textit{``How is electricity like flowing water?''}. LLMs are qualitatively evaluated on their ability to provide a convincing answer.

\subsection{Spatial Reasoning}
A small experimental dataset is constructed by taking cues from SpartQA \cite{spartqa}. Additionally, the model's spatial reasoning capabilities are also evaluated using a comprehensive task with follow up questions in a free-form conversational language format. As before, GPT-3 \textit{davinici-003} is used for experiments with the controlled dataset and ChatGPT is used for conversational style dataset with follow up questions.
\subsubsection{Controlled Dataset}
\begin{figure}[]
            % \centering
             \includegraphics[width=.5\textwidth]{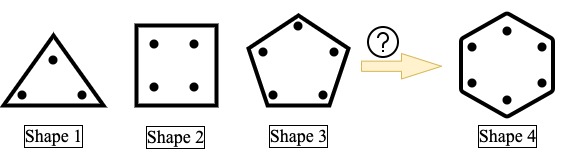}
            \caption{Spatial Reasoning:$\;$Visualization of Controlled Dataset Example}
        \label{fig:SR1}
\end{figure}

The spatial reasoning evaluation task is designed as a textual entailment task using a controlled dataset, which is constructed in-house following inspiration from SpartQA. In this task, the model is presented with a textual description of three objects and then asked to predict whether a fourth described object follows closely to the pattern of the three previous shapes. The model's response is constrained to a True or False output which provides a clear metric for evaluation. Figure \ref{fig:SR1} provides a visual example of the textually described problem. The model's accuracy is evaluated by the number of correct predictions.

\subsubsection{Conversational Prompts}
This subset of experiments has been inspired by human cognitive abilities and their capability to understand and navigate their environment by constructing a mental map. The models are presented with a textual description of a house layout consisting of four rooms and a hallway. Thereafter, they are probed with a series of questions to assess their ability to reason spatially.

\subsection{Ethical and Moral Reasoning}
Ethics and morality is a unique feature of humanity. It's interpretation varies substantially across individuals and societies, and thus lacks universal consensus. To evaluate the moral reasoning capabilities of ChatGPT, I adopt a strategy of asking open-ended, free-form questions in natural language to elicit responses to morally ambiguous scenarios. I then press the model to take a definitive stance in challenging dilemmas to explore its understanding of moral nuances and ambiguities. As the results are not easily quantifiable, they are left open to qualitative interpretation by the readers of this paper.

\section{Experimental Results and Analysis}
\subsection{Analogical Reasoning}
\subsubsection{Controlled Dataset}
\begin{table*}[]
\small
\begin{tabular}{llccc}
\hline
Relation          & Prompt                                                  & \begin{tabular}[c]{@{}c@{}}Correct \\ Answers\end{tabular} & \begin{tabular}[c]{@{}c@{}}Predicted\\ Answers\end{tabular} & Correct \\ \hline
\textbf{Hypernym} & \textit{Velociraptor} is to \textit{dinosaur} as \textit{duck} is to \underline{$\;\;\;\;\;$}.       & fowl, bird, vertebrate...                                  & bird                                                        & \cmark     \\
\textbf{Hypernym} & \textit{Sofa} is to \textit{piece of furniture} as \textit{stapeler} is to \underline{$\;\;\;\;\;$}. & device, machine, entity...                                 & office supply                                               & \xmark      \\ \hline
\textbf{Hyponym}  & \textit{Citrus} is to \textit{tangerine} as \textit{dessert} is to \underline{$\;\;\;\;\;$}.         & fruit, cake, ice-cream...                                  & cake                                                        & \cmark     \\
\textbf{Hyponym}  & \textit{Season} is to \textit{summertime} as \textit{month} is to \underline{$\;\;\;\;\;$}.          & january, february, march...                                & winter                                                      & \xmark       \\ \hline
\textbf{Meronym}  & \textit{Secretary} is to \textit{staff} as \textit{galaxy} is to \underline{$\;\;\;\;\;$}.           & universe                                                   & universe                                                    & \cmark    \\
\textbf{Meronym}  & \textit{Wolf} is to \textit{pack} as \textit{letter} is to \underline{$\;\;\;\;\;$}.                 & alphabet                                                   & envelope                                                    & \xmark      \\ \hline
\textbf{Synonym}  & \textit{Monkey} is to \textit{gorilla} as \textit{necessary} is to \underline{$\;\;\;\;\;$}.         & essential, vital, required...                              & essential                                                   & \cmark     \\
\textbf{Synonym}  & \textit{Well} is to \textit{flourishing} as \textit{nap} is to \underline{$\;\;\;\;\;$}.             & sleep, slumber                                             & refreshed                                                   & \xmark       \\ \hline
\textbf{Antonym}  & \textit{Young} is to \textit{hoary} as \textit{generous} is to \underline{$\;\;\;\;\;$}.             & stingy, beggarly, mean...                                  & stingy                                                      & \cmark     \\
\textbf{Antonym}  & \textit{Dangerous} is to \textit{harmless} as \textit{noisy} is to \underline{$\;\;\;\;\;$}.         & silent, dumb, mute...                                      & quiet                                                       & \xmark       \\ \hline
\end{tabular}
\caption{GPT-3 Predictions on Lexicographic Semantic Analogies - BATS Dataset}
\label{table: 1}
\end{table*}

Table \ref{table: 1} presents some notable success and failure cases of the GPT-3 model on the curated dataset of 200 examples for the analogical reasoning task. The model achieves an exact-match accuracy of 53\%. Despite a performance which may appear as suboptimal, upon deeper analysis of the failure cases it becomes apparent in the absence of contextual cues and answer choices, it may pose a significant challenge even to human reasoning. This observation highlights the importance of qualitative evaluation of LLMs on free-form natural language questions such as those which maybe used in colloquial discourse.
\subsubsection{Conversational Prompts}
This section presents results of evaluation of ChatGPT's analogical reasoning abilities, a hallmark of human cognition that enables us to draw inferences and make predictions by identifying similarities between seemingly dissimilar concepts. Analogical reasoning is also demonstrated by us by our ability to explain complex abstract ideas using simpler, relatable examples. To test ChatGPT's performance in this area, I prompt it to explain a complex concept with the help of a simpler one and provide examples in Appendix \ref{Appendix: A}. The examples demonstrate the model's impressive ability to utilize analogies to simplify complex concepts. However, it is noteworthy that the model tends to justify incorrect claims instead of rejecting them, as evidenced in Example 1 and 2. The latter example is particularly striking as the model produces an inaccurate description of \textit{how electricity is generated} to keep the analogy running. These results point out the strengths of the model in analogical reasoning as well as their shortcomings.
\subsection{Spatial Reasoning}
\subsubsection{Controlled Dataset}
 \begin{figure}[]
            % \centering
             \includegraphics[width=.5\textwidth]{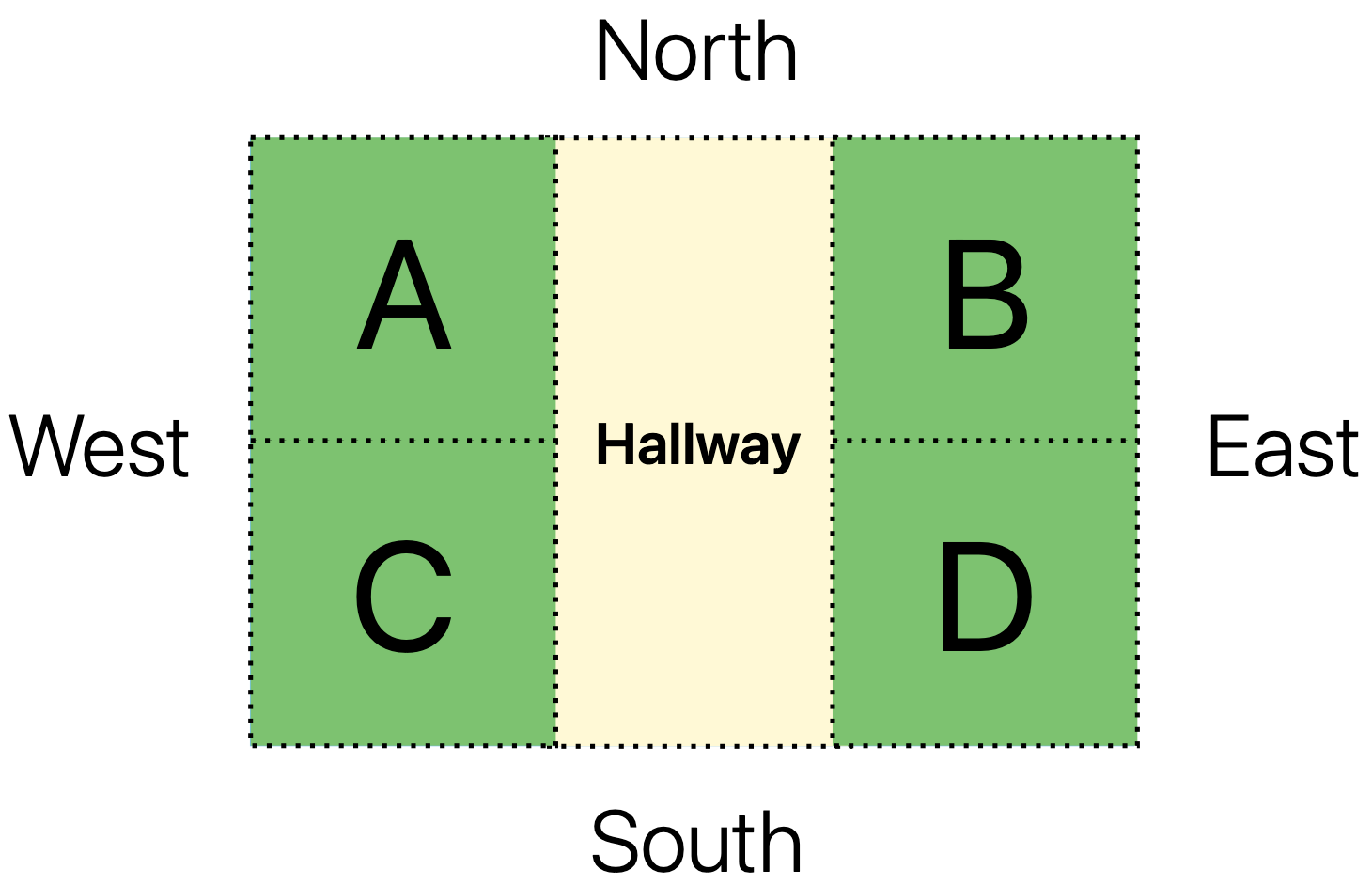}
            \caption{Spatial Reasoning:$\;$Visualization of the layout of house described via textual description}
        \label{fig:SR2}
\end{figure}
The spatial reasoning abilities of GPT-3 are evaluated through a manually curated controlled dataset with a specific instruction prompt (Appendix \ref{Appendix: B}), challenging the model to determine whether a particular shape continues a pattern established by three previous shapes. The task description prompt is followed by a textual description of four shapes, suffixed with a ``True'' or ``False'' answer request. Table \ref{table: 2} outlines the various experiments with the textual descriptions for the different shapes. The model displays an accuracy of 55\%, which is only slightly above chance. The model exhibits a strong inclination of predicting \textit{True}, and only predicting \textit{False} when the task is made relatively simple with the inconsistency in shape 4 being made glaringly obvious. From the results it can be seen that GPT-3 pays more attention to the numbers than the colors and shapes while determining its prediction. This highlights the limitations of LLMs when it comes to spatial reasoning based solely on textual description of objects.
\subsubsection{Conversational Prompts}
To evaluate the spatial reasoning skills of LLMs in a more realistic and human-like manner, a scenario based on real-life spatial challenges was constructed. A detailed textual description of a house with four rooms and one hallway was provided to the model to visualize. Figure \ref{fig:SR2} illustrates the problem statement. Impressively, as shown in Figure \ref{fig:SR3} the model accurately modeled the house based on the description. Furthermore, Appendix \ref{Appendix: B} showcases that the model is capable of accurately reasoning about spatial and positional information, but only if the information can be directly inferred from the provided context. However, when presented with queries that require multi-step spatial reasoning, and where significant deviation from the written context is needed, the model struggles to provide correct answers and resorts to supporting its reasoning with information that is hallucinated. This lends further support to the point that, spatial reasoning problems which can be solved by humans with ease, presents a significant challenge to LLMs. The results suggest that the model's spatial reasoning skills are severely limited due to its lack of visual information during training.
\subsection{Ethical and Moral Reasoning}
The evaluation of LLMs on moral reasoning is a particularly complex and fascinating problem due to a lack of universal consensus on one correct answer. Nonetheless, the examples presented in Appendix \ref{Appendix: C} demonstrate that the model's remarkable ability to decipher the nuances in morally ambiguous situations, and proposes solutions which often exceed my expectations. However, given the inherent subjectivity of morality, I leave the model's results on this task open to interpretation by the reader.
\section{Conclusion}
In conclusion, this paper presented an evaluation of the reasoning capabilities of LLMs, particularly focusing on the domains of analogical, spatial and moral reasoning. Our experiments showed that while LLMs perform moderately well on tasks pertaining to analogical reasoning, their performance in tasks requiring spatial reasoning is not up to the mark. Moreover, the subjective nature of morality makes it challenging to evaluate their moral reasoning abilities. Nonetheless, LLMs showcase moral reasoning abilities which exceed expectations.\\
Overall, the evaluation of LLMs on different domains of reasoning tasks is an important step towards understanding their limitations, and can better inform their potential applications in real-world.

\clearpage

\bibliography{naaclhlt2016}
\bibliographystyle{naaclhlt2016}
\clearpage

\appendix
\section*{Appendix}
\section{Analogical Reasoning Examples\label{Appendix: A}}
The following prompt is provided as context before each example to condition ChatGPT in providing brief answers:\\
\textbf{I am going to ask you a question about explaining a difficult concept using the examples of simpler and more relatable concepts. Please answer it as clearly as you can while keeping it brief. }\\

\noindent\textbf{Example 1: \label{Ex: A1}}\\
\textbf{Author}: Can you explain how a camera works by taking the example of a human eye? \\
\textbf{ChatGPT}: A camera works in a similar way to the human eye. Just as the eye has a lens that focuses light onto the retina at the back of the eye, a camera has a lens that focuses light onto a digital sensor or film. The amount of light that enters the camera is controlled by the aperture, which is like the pupil of the eye.

In both cases, the light-sensitive surface (the retina in the eye and the digital sensor or film in the camera) captures the image that is projected onto it. The captured image is then processed by the brain in the case of the eye, or by a computer in the case of a digital camera, to produce the final image.\\

\noindent\textbf{Author}: Then how can a camera zoom in and see images very far away but humans can't do the same?\\
\textbf{ChatGPT}: Cameras can zoom in and see faraway objects by using lenses with a variable focal length that allow them to adjust the magnification of the image. When a camera zooms in, the lens changes shape to increase the focal length, which magnifies the image and makes it appear closer. Some cameras also have telephoto lenses that further increase the magnification.

In contrast, the human eye has a fixed focal length and cannot change its shape to adjust the magnification of the image. This means that when we look at distant objects, they appear smaller and less detailed than they do through a camera with zooming capabilities. However, the human eye has other advantages over cameras, such as a wider field of view and the ability to see in a range of lighting conditions.
\\

\noindent[Attempt to trick the model to draw false analogies:]\\
\noindent\textbf{Author:} How can cameras see very well in the dark but humans aren't able to do so?\\
\textbf{ChatGPT:} Cameras can see well in the dark because they have a sensor that is more sensitive to light than the human eye. This means that they can capture more light and produce a brighter image in low light conditions.

In addition, many modern cameras have a feature called "night vision" or "infrared" mode that can capture images in very low light conditions by using infrared light. Infrared light is outside the visible spectrum of light that the human eye can detect, but cameras can capture it using a specialized sensor.

In contrast, the human eye has limited sensitivity to light in low light conditions, and cannot detect infrared light. However, the eye can adjust to low light conditions over time by dilating the pupils to allow more light in and by adjusting the sensitivity of the retina, a process known as dark adaptation.\\

\noindent\textbf{Author:} How can humans see very well in the dark but cameras aren't able to do the same?\\
\textbf{ChatGPT:} Humans can see well in the dark because the human eye has a special layer of cells in the retina called rods that are highly sensitive to light. These rods allow the eye to detect very low levels of light, which is why we can see in low light conditions.

In addition, the human eye has the ability to adapt to changes in lighting conditions by adjusting the size of the pupil, which controls the amount of light that enters the eye. This adaptation process allows the eye to function well in a wide range of lighting conditions, from bright sunlight to complete darkness.

However, cameras are limited in their ability to capture images in the dark because they rely on the amount of available light to produce an image. While cameras can use features like high ISO settings or long exposures to capture images in low light conditions, there is a limit to how much light can be captured before the image becomes too noisy or blurry. This is why cameras often struggle to produce clear images in very dark conditions.\\
\clearpage

%A. EXAMPLE 2

\noindent\textbf{Example 2: \label{Ex: A2}}\\
\textbf{Author}: Can you explain how electricity works by taking the example of water?\\
\textbf{ChatGPT}: A useful analogy for understanding electricity is to think of it as water flowing through a system of pipes. In this analogy, the water represents the flow of electric charge, which is what makes electricity work.

Just like water, electric charge can flow through a conductor (a material that allows the flow of electric charge), like a wire. When an electrical current flows through a wire, it's like water flowing through a pipe.

The amount of electric charge that flows through a conductor is determined by the voltage, which is like the pressure of the water in the pipes. A higher voltage means that more electric charge is flowing through the conductor, just as higher water pressure means that more water is flowing through the pipes.

Similarly, the resistance of a conductor is like the size of the pipe that the water is flowing through. A smaller pipe will restrict the flow of water, just as a higher resistance in a conductor will restrict the flow of electric charge.
Just as a water turbine can be used to generate electricity by using the pressure of the water to turn a generator, {\color{red} electric generators use the flow of electric charge to create a magnetic field that turns a rotor and generates electrical energy.}

In summary, the flow of electric charge in a circuit can be understood through the analogy of water flowing through pipes. Voltage is like the pressure of the water, resistance is like the size of the pipe, and {\color{red}electric generators work like water turbines, using the flow of electric charge to generate electrical energy}.\\

\noindent[ChatGPT's generates factually incorrect information in a very convincing way to keep the analogy running. I try to highlight the incorrect facts with another question.]\\
\noindent\textbf{Author}: Don't electric generators convert mechanical energy from an external source into electrical energy?\\
\textbf{ChatGPT}: Yes, you're absolutely right! I apologize for the mistake in my previous response. Electric generators do indeed convert mechanical energy from an external source into electrical energy, rather than using the flow of electric charge to generate electrical energy as I previously stated.

In an electric generator, a magnet is rotated inside a coil of wire, which generates an electrical current in the wire. This is similar to how a water turbine is used to generate electricity by using the kinetic energy of the moving water to turn a rotor and generate electrical energy.

So to revise the analogy, just as a water turbine can be used to generate electricity by using the kinetic energy of the moving water to turn a generator, electric generators use the mechanical energy from an external source to turn a rotor and generate electrical energy. The flow of electric charge in a circuit is still analogous to water flowing through pipes, with voltage representing the pressure of the water and resistance representing the size of the pipe.\\

\noindent[Trying to get ChatGPT to commit the same mistake.]\\
\noindent\textbf{Author}:But I thought electric generators use electric charge (electricity) to generate electricity?\\
\textbf{ChatGPT}: Electric generators actually use mechanical energy to generate electricity. This mechanical energy can come from a variety of sources, such as the steam produced by burning fossil fuels or nuclear reactions, the kinetic energy of wind or water, or the rotational energy of turbines powered by falling water or steam.

Inside an electric generator, this mechanical energy is used to turn a rotor, which is surrounded by a stationary coil of wire. As the rotor turns, it creates a changing magnetic field that induces an electric current in the wire coil through a process called electromagnetic induction. This induced current is the electrical energy that is generated by the generator.

So in summary, electric generators do not use electricity to generate more electricity. Instead, they use mechanical energy to generate electrical energy through electromagnetic induction. The flow of electric charge in a circuit can still be compared to the flow of water through pipes, but it is not directly involved in the process of generating electrical energy in an electric generator.
\clearpage
\noindent \textbf{Example 3 \label{Ex: A3}}:\\
\noindent[This example from Shakespeare's popular play Macbeth is intended to evaluate whether ChatGPT can draw analogies to explain abstract concepts.]\\
\textbf{Author}: ``Life’s but a walking shadow, a poor player. 
That struts and frets his hour upon the stage
And then is heard no more. It is a tale 
Told by an idiot, full of sound and fury,
Signifying nothing.''
Can you explain how life is like a walking shadow?\\
\textbf{ChatGPT}: Certainly! The quote you provided is from William Shakespeare's play Macbeth, where the character Macbeth is contemplating the fleeting nature of life.

When he says "life's but a walking shadow," he is comparing life to a shadow, which is an image or a representation of something else. A shadow is not a tangible object, but rather a temporary projection of something that exists. Similarly, life is also temporary and fleeting, just like a shadow.

Furthermore, when he says "a poor player, that struts and frets his hour upon the stage," he is comparing life to a performance on a stage. Just like a play, life is temporary and limited. We only have a certain amount of time to play our part in the grand scheme of things, and then our time on stage is over.

Finally, when Macbeth says that life is "full of sound and fury, signifying nothing," he is suggesting that life is filled with meaningless noise and drama. We may spend our time pursuing various goals and ambitions, but in the grand scheme of things, it may not matter much at all. Our actions and accomplishments may be forgotten or irrelevant once we are gone, just like a tale told by an idiot.

So, in summary, Macbeth is using the metaphor of a walking shadow to describe how life is fleeting and temporary, and the metaphor of a poor player on a stage to emphasize how we have a limited time to play our part. Ultimately, he suggests that the noise and drama of life may be meaningless in the grand scheme of things.\\

\noindent\textbf{Author}: But unlike a shadow which leaves no trace after it vanishes, the tales of a life continue on for ages. So I think this analogy doesn't hold. How do you then justify then author's claims?\\
\textbf{ChatGPT}: You make a valid point. While a shadow may disappear without leaving any trace, a person's life often has a lasting impact on the world, and their story may be told and remembered for generations.

However, I believe that Shakespeare's comparison of life to a walking shadow is not intended to be taken literally, but rather as a poetic expression of the transience of human existence. The comparison is a way of emphasizing the fleeting nature of life and the fact that it can be gone in an instant, just like a shadow that disappears with the slightest change in light or movement.

Similarly, the comparison of life to a tale told by an idiot suggests that, while a person's life may have significance to them, it may not have any broader meaning in the grand scheme of things. The noise and fury of life may be all-consuming to the individual, but in the larger context of the universe, they may not signify anything of importance.

In summary, while the analogy may not be a perfect match for the way that a person's life is remembered, it is a powerful and evocative metaphor that captures the fleeting and transitory nature of human existence, as well as the potential lack of significance of individual actions in the larger context of the universe.

\clearpage
\section{Spatial Reasoning Examples \label{Appendix: B}}

\begin{table*}[]
\small
\begin{tabular}{lllllccc}
\hline
\multicolumn{1}{c}{Pattern}                                                                                                                                                                 & \multicolumn{1}{c}{Shape 1}                                         & \multicolumn{1}{c}{Shape 2}                                         & \multicolumn{1}{c}{Shape 3}                                         & \multicolumn{1}{c}{Shape 4}                                          & \begin{tabular}[c]{@{}c@{}}Correct\\ Answer\end{tabular} & \begin{tabular}[c]{@{}c@{}}Predicted\\ Answer\end{tabular} & Correct \\ \hline
\multirow{2}{*}{\begin{tabular}[c]{@{}l@{}}Number of sides\\ in the shapes increase \\ by one\end{tabular}}                                                                                 & Triangle                                                            & Square                                                              & Pentagon                                                            & Hexagon                                                              & True                                                     & True                                                       & \cmark   \\  \\
                                                                                                                                                                                            & Triangle                                                            & Square                                                              & Pentagon                                                            & Heptagon                                                             & False                                                    & True                                                       & \xmark      \\ \hline
\multirow{2}{*}{\begin{tabular}[c]{@{}l@{}}Number of sides \\ in the shapes increase \\ by one and number of \\ dots in the shape increase \\ by one\end{tabular}}                          & \begin{tabular}[c]{@{}l@{}}Triangle with \\ 2 dots\end{tabular}     & \begin{tabular}[c]{@{}l@{}}Square with \\ 3 dots\end{tabular}       & \begin{tabular}[c]{@{}l@{}}Pentagon with \\ 4 dots\end{tabular}     & \begin{tabular}[c]{@{}l@{}}Hexagon with \\ 5 dots\end{tabular}       & True                                                     & True                                                       & \cmark \\   \\ 
                                                                                                                                                                                            & \begin{tabular}[c]{@{}l@{}}Triangle with \\ 2 dots\end{tabular}     & \begin{tabular}[c]{@{}l@{}}Square with \\ 3 dots\end{tabular}       & \begin{tabular}[c]{@{}l@{}}Pentagon with \\ 4 dots\end{tabular}     & \begin{tabular}[c]{@{}l@{}}Heptagon with \\ 5 dots\end{tabular}      & False                                                    & True                                                       & \xmark \\      \\ \hline
\multirow{2}{*}{\begin{tabular}[c]{@{}l@{}}The color of dots \\ alternates between \\ red and black\end{tabular}}                                                                           & \begin{tabular}[c]{@{}l@{}}Square with 1\\ red dot\end{tabular}     & \begin{tabular}[c]{@{}l@{}}Square with \\ 1 black dot\end{tabular}  & \begin{tabular}[c]{@{}l@{}}Square with \\ 1 red dot\end{tabular}    & \begin{tabular}[c]{@{}l@{}}Square with \\ 1 black dots\end{tabular}  & True                                                     & True                                                       & \cmark     \\ \\
                                                                                                                                                                                            & \begin{tabular}[c]{@{}l@{}}Square with \\ 1 red dot\end{tabular}    & \begin{tabular}[c]{@{}l@{}}Square with 1 \\ black dot\end{tabular}  & \begin{tabular}[c]{@{}l@{}}Square with \\ 1 red dots\end{tabular}   & \begin{tabular}[c]{@{}l@{}}Square with \\ 1 red dots\end{tabular}    & False                                                    & True                                                       & \xmark      \\ \hline
\multirow{2}{*}{\begin{tabular}[c]{@{}l@{}}The color of dots \\alternates between red \\ and black and, the \\ number of dots is \\ equal to the number \\of sides  of the shape\end{tabular}} & \begin{tabular}[c]{@{}l@{}}Triangle with \\ 3 red dots\end{tabular} & \begin{tabular}[c]{@{}l@{}}Square with \\ 4 black dots\end{tabular} & \begin{tabular}[c]{@{}l@{}}Pentagon with \\ 5 red dots\end{tabular} & \begin{tabular}[c]{@{}l@{}}Hexagon with \\ 6 red dots\end{tabular}   & False                                                    & True                                                       & \xmark  \\    \\
                                                                                                                                                                                            & \begin{tabular}[c]{@{}l@{}}Triangle with \\ 3 red dots\end{tabular} & \begin{tabular}[c]{@{}l@{}}Square with \\ 4 black dots\end{tabular} & \begin{tabular}[c]{@{}l@{}}Pentagon with \\ 5 red dots\end{tabular} & \begin{tabular}[c]{@{}l@{}}Hexagon with \\ 600 red dots\end{tabular} & False                                                    & False                                                      & \cmark \\     \\\\ \hline
\end{tabular}
\caption{GPT-3 Performance on textually described spatial reasoning tasks}
\label{table: 2}
\end{table*}

\subsection{Conversational Prompts}
The layout of the house is described via the following prompt:\\
\textbf{Imagine a house with a rectangular shape consisting of 4 square rooms of equal size and a spacious rectangular hall stretching from the North to the South direction. The house is designed in such a way that Room A occupies the North-West corner, Room C occupies the South-West corner, Room B is located in the North-East corner, and Room D is situated in the South-East corner. The rectangular hall serves as a central pathway that separates the house into two sections, with rooms A and C on the West and rooms B and D on the east of the hall. Notably, the rooms on either side of the hall share a common wall.}\\

\noindent\textbf{Author}: Did you understand the layout of the house?\\
\textbf{ChatGPT}: Yes, I understand the layout of the house you described. The house has a rectangular shape and is divided into four equal-sized square rooms, with Room A in the North-West corner, Room C in the South-West corner, Room B in the North-East corner, and Room D in the South-East corner. The rectangular hall runs from the North to the South direction and separates the house into two sections, with rooms A and C on the West and rooms B and D on the East of the hall. The rooms on either side of the hall share a common wall.\\

\begin{figure}[]
            % \centering
            \includegraphics[keepaspectratio, width=0.5\textwidth]{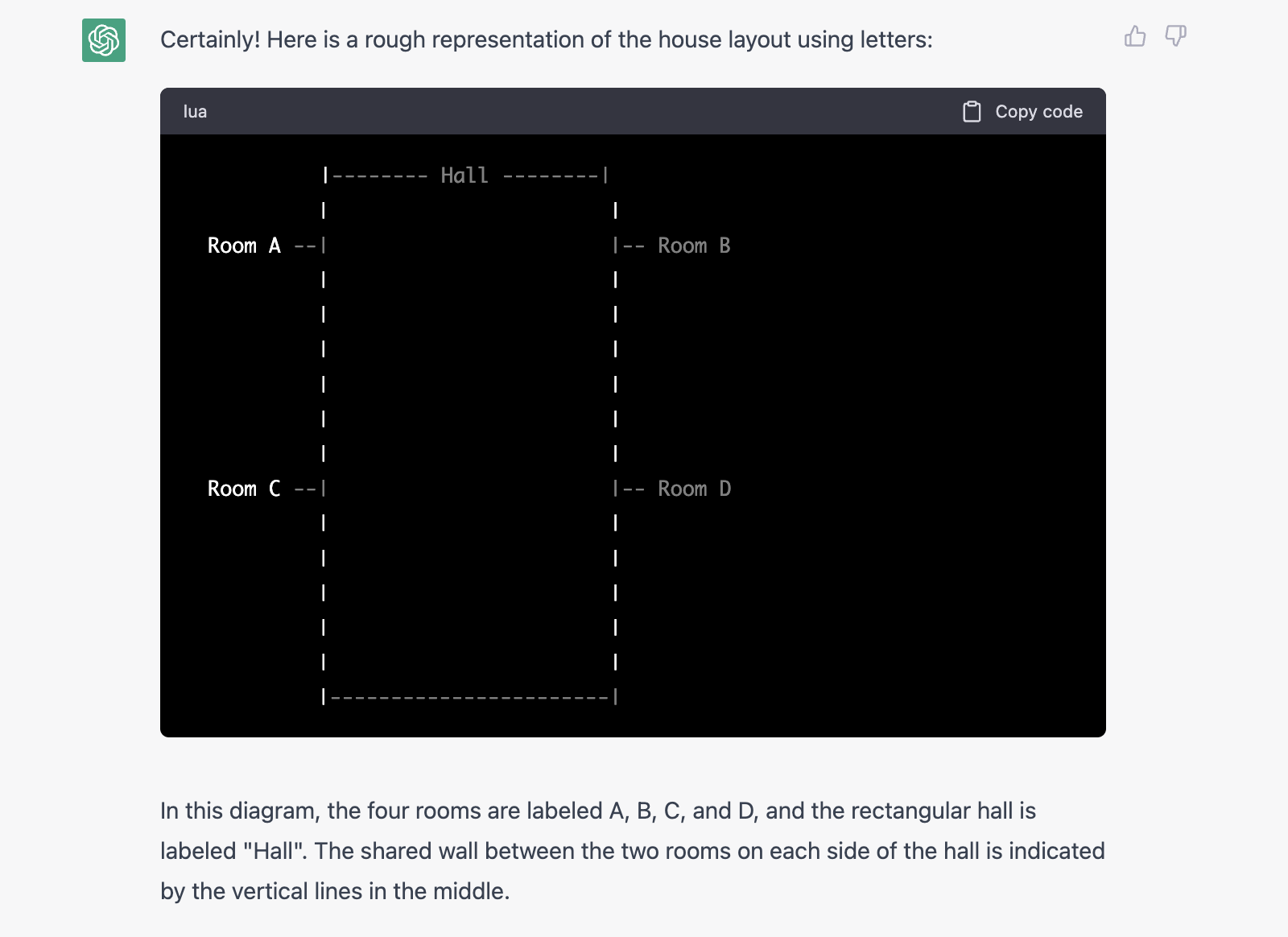}
            \caption{Spatial Reasoning:$\;$Visualization of the layout of house as imagined by ChatGPT \label{fig:SR3}}
\end{figure}

\noindent\textbf{Author}: Can you draw a diagram of the layout of the house using letters and symbols?\\
\textbf{ChatGPT}: \textit{Replies as shown in Figure} \ref{fig:SR3}\\
            
\noindent\textbf{Author}: Great, now I'll ask you a series of questions. Please answer them on the basis of your understanding of the layout of the house.\\
\textbf{ChatGPT}: Sure, I'll do my best to answer your questions based on my understanding of the house layout.\\

\noindent\textbf{Author}: Suppose I'm sitting in Room A and I hear someone calling me from the adjoining room. The person who is calling me is sitting in which room?\\
\textbf{ChatGPT}: If you are sitting in Room A and you hear someone calling you from the adjoining room, then the person who is calling you is most likely sitting in Room C, as Room C is located to the south of Room A and is the only adjoining room on that side of the hall. $\;\;$[\cmark]\\

\noindent\textbf{Author}: Suppose I'm sitting in Room A and I hear some noise from the room directly opposite to me across the hallway. Which room is the noise coming from?\\
\textbf{ChatGPT}: If you are sitting in Room A and hear some noise from the room directly opposite to you across the hallway, then the noise is most likely coming from Room B, as Room B is located directly across the hallway from Room A. $\;\;$[\cmark]\\ 

\noindent[Question to evaluate the ability to shift the mental compass and accordingly give directions relative to the shifted axis.]\\
\noindent\textbf{Author}: Suppose I'm watching TV in the hall. The TV is mounted on the wall shared by rooms A and C. I hear some sound from the room behind me. Which rooms can the sound be coming from?\\
\textbf{ChatGPT}: If you are watching TV in the hall and the TV is mounted on the wall shared by Rooms A and C, and you hear some sound from the room behind you, then the sound could be coming from either Room B or Room D, as these are the two rooms located behind you across the hallway. $\;\;$[\cmark] \\

\noindent\textbf{Author}: Suppose I'm watching TV in the hall. The TV is mounted on the wall shared by rooms A and C. I hear some sound from the room behind me. Which rooms can the sound be coming from?\\
\textbf{ChatGPT}: If you are watching TV in the hall and the TV is mounted on the wall shared by Rooms A and C, and you hear some sound from the room behind you, then the sound could be coming from either Room B or Room D, as these are the two rooms located behind you across the hallway. $\;\;$[\cmark] \\

\noindent\textbf{Author}: I am sitting in Room C. Now if I get out of the room and walk to the room diagonally across from me, which room will I arrive in?\\
\textbf{ChatGPT}: If you are sitting in Room C and you get out of the room and walk diagonally across from you, you will arrive in Room B. This is because Rooms B and C are diagonally opposite to each other in the rectangular layout of the house. $\;\;$[\cmark] \\

\noindent\textbf{Author}: Now will the TV share a wall with the room I'm currently in?\\
\textbf{ChatGPT}: No, if you are in Room C and you walk diagonally across to Room B, the TV will not share a wall with the room you are currently in. The TV is mounted on the wall that is shared by Rooms A and C, which are on the other side of the hall from Rooms B and D. Therefore, the TV will not share a wall with Room B or any other room on the east side of the hall. $\;\;$[\cmark] \\

\noindent[Moving to questions which have no mention in the previous context and can be soled solely by having a sense of directions and spatial awareness.]
\noindent\textbf{Author}: I hear someone enter the house from the door located on the Southern end of the hall and go straight to the room on the other end of the hall to their left side. Which room did they go to?\\
\textbf{ChatGPT}: If someone enters the house from the door located on the Southern end of the hall and goes straight to the room on the other end of the hall to their left side, then they went to Room B. This is because Room B is located on the left-hand side of the hall when someone enters the house from the door on the Southern end of the hall, and it is the room on the other end of the hall from that door. $\;\;$[\xmark] \\
\clearpage
\noindent[Pointing out the error made by ChatGPT to ensure it's mental image is reflecting the current state.]
\noindent\textbf{Author}: That's incorrect. The person entered room A.\\
\textbf{ChatGPT}: I apologize for the mistake in my previous response. You are correct that if someone enters the house from the door located on the Southern end of the hall and goes straight to the room on the other end of the hall to their left side, then they would actually be entering Room A. This is because Room A is the room on the left-hand side of the hall when someone enters the house from the door on the Southern end of the hall, and it is the room on the other end of the hall from that door. $\;\;$ \\

\noindent\textbf{Author}: Considering all the other rooms in the house are empty except for the room I'm currently in and the room occupied by the person who just came. Which rooms are empty?\\
\textbf{ChatGPT}: If all the other rooms in the house are empty except for the room you are currently in and the room occupied by the person who just entered, then Rooms B, C, and D are empty. This is because you have already identified that the person entered Room A, and the other three rooms are on the opposite side of the hallway and are not adjacent to Room A. Therefore, they must be empty. $\;\;$[\xmark] \\

\noindent[Upon pointing out the error, ChatGPT begins to correct its answer but at the same explaining the new answer with factually incorrect reasoning.]\\
\noindent\textbf{Author}: That's incorrect as I'm currently in Room B. So Room B is not empty as well.\\
\textbf{ChatGPT}: I apologize for the mistake in my previous response. If you are currently in Room B, then Rooms C and D are empty, while Rooms A and B are occupied by you and the person who entered the house, respectively. This is because Rooms A and B are on the same side of the hallway as Room B, while Rooms C and D are on the opposite side of the hallway and are not adjacent to Room B. Therefore, they must be empty.$\;\;$[\xmark] \\

\noindent[ChatGPT retains spatial information stored in its context.]\\
\noindent\textbf{Author}: The person in Room A gets up to leave the house. Where will they exit from considering there is only one entrance in the house?\\
\textbf{ChatGPT}: If the person in Room A gets up to leave the house, they will exit from the door located on the Southern end of the hall. This is the only entrance in the house, so it is the only way for the person to leave.$\;\;$[\cmark] \\

\noindent[ChatGPT is able to reason about the relative positions of the room.]\\
\noindent\textbf{Author}: I get up and head South. Which room will I arrive in?\\
\textbf{ChatGPT}: If you get up and head South from Room B, you will arrive in Room D. Room D is the southernmost room in the house, and it is located on the same side of the hallway as Room B. Therefore, if you head South from Room B, you will arrive in Room D.$\;\;$[\cmark] \\

\noindent[Question meant to mislead the model in giving incorrect answer.]\\
\noindent\textbf{Author}: I hear some sound coming from East. Which room is the sound coming from?\\
\textbf{ChatGPT}: If you hear some sound coming from East, the sound is coming from either Room B or Room D. These are the two rooms located on the East side of the hallway. Without any additional information about the specific location of the sound, it is impossible to determine which room it is coming from.$\;\;$[\xmark] \\

\noindent\textbf{Author}: I hear some sound coming from the TV. From which direction relative to me is the sound coming from?\\
\textbf{ChatGPT}: If you hear some sound coming from the TV, the sound is coming from the direction of the wall that is shared by Rooms A and C. This is because the TV is mounted on the wall shared by these two rooms, and the sound is likely to be coming from the speakers of the TV, which are directed towards the viewers. Therefore, the sound is coming from the direction of the wall that is shared by Rooms A and C, which is on the West side of the hallway relative to Rooms B and D.$\;\;$[\cmark] \\

\noindent\textbf{Author}: So will it be from the left or right side relative to me?\\
\textbf{ChatGPT}: If you are sitting in Room B, and the TV is mounted on the wall shared by Rooms A and C, the sound is coming from your left side relative to your position in Room B. This is because Room A is located to the left of Room B, and the TV is mounted on the wall shared by Rooms A and C. Therefore, the sound is likely to be coming from the left side of your position in Room B.$\;\;$[\cmark] \\

\noindent[Multifaceted question aimed to analyze the reasoning of model over local position as well as the mental map of the environment.]\\
\noindent\textbf{Author}: Now I walk over to room directly opposite to me across the hallway. From which direction is the sound coming from now?\\
\textbf{ChatGPT}: If you walk over to the room directly opposite to Room B across the hallway, which is Room D, the sound from the TV is now coming from your right side relative to your position in Room D. This is because Room C is located to the right of Room D, and the TV is mounted on the wall shared by Rooms A and C. Therefore, the sound is likely to be coming from the right side of your position in Room D.$\;\;$[\xmark] \\

\noindent[The model hallucinates the answer and also loses context of our position.]

\section{Moral Reasoning \label{Appendix: C}}
\noindent \textbf{Example1}\\
\textbf{Author}:Do you think it is ever morally acceptable to download copyrighted material without paying for it? Pick one side and provide a brief answer to justify your position.\\
\textbf{ChatGPT}:As an AI language model, I do not have personal opinions or beliefs, but I can present both sides of the argument.

On one hand, some people argue that downloading copyrighted material without paying for it is never morally acceptable because it is a form of theft. They argue that creators deserve to be compensated for their work and that downloading copyrighted material without paying for it is equivalent to stealing.

On the other hand, some people argue that there are circumstances where it can be morally acceptable to download copyrighted material without paying for it. For example, they may argue that if the material is not available in the downloader's country or is prohibitively expensive, then it may be acceptable to download it without paying for it. Additionally, they may argue that if the material is old and no longer being actively marketed, then it may be acceptable to download it without paying for it.

However, it's worth noting that the law does not generally allow for these exceptions. In most countries, downloading copyrighted material without permission or payment is illegal, regardless of the downloader's intentions or circumstances.\\

\noindent\textbf{Author}:Suppose a student needs to urgently download a book to study for their final exam and does not have the time or the funds to go purchase it from the bookstore. Assume, that the decision falls upon you whether to allow it or not. What would you do?\\
\textbf{ChatGPT}:As an AI language model, I cannot make decisions or take actions. However, I can provide some information that may be helpful in making a decision in this scenario.

In general, it is not legal to download copyrighted material without permission or payment, even in cases where the downloader feels they have a legitimate reason to do so. Urgency and financial constraints do not typically excuse copyright infringement.

That being said, there are some alternatives to consider in this situation. For example, the student may be able to access the book through a library, either by checking it out physically or accessing an electronic copy through the library's online database. Additionally, the student may be able to purchase or rent an electronic copy of the book at a lower cost than a physical copy.

Ultimately, it is important to consider both the legal and ethical implications of downloading copyrighted material without permission or payment. While urgency and financial constraints may make it tempting to take this action, it is important to consider alternative options that are both legal and ethical.\\

\noindent\textbf{Author}: Suppose the student is from a financially deprived background, and them performing well in the exams and securing a good job can create a huge difference in the life of their family. Do you think it is moral in this case to download a copyrighted book?\\
\textbf{ChatGPT}: As an AI language model, I cannot provide a straightforward answer to this question as it involves subjective moral and ethical considerations.

However, it is important to note that legality and morality are not always the same thing. While downloading copyrighted material without permission or payment is illegal, some people may argue that there are circumstances where it can be morally justifiable.

In the scenario you've presented, the student's financial situation and the potential impact of their exam performance on their family could be seen as mitigating factors that may make it more understandable why they would choose to download a copyrighted book without paying for it.

That being said, it's important to consider the potential consequences of such actions, both for the individual and for society as a whole. It's possible that downloading copyrighted material without permission or payment could result in legal and financial penalties for the individual, which could exacerbate their already difficult financial situation. Additionally, if widespread copyright infringement becomes normalized, it could have negative implications for the creators and industries that rely on copyright protection to earn a living.

Ultimately, it's up to individuals to weigh the legal and ethical considerations of their actions and make a decision that they feel is appropriate. However, it's important to be aware of the potential consequences of those actions and to consider alternative options that are both legal and ethical.\\

\noindent\textbf{Author}: Now consider, that the book the student wants to purchase is by a struggling author and people pirating his book could put him out of a job and plunge him into a life of despair. Do you think it would be moral for the student to download the copyrighted book then?\\
\textbf{ChatGPT}: As an AI language model, I cannot provide a straightforward answer to this question as it involves subjective moral and ethical considerations.

However, it is important to consider the potential impact of one's actions on the creator of the copyrighted material. If the author of the book is struggling financially and relies on sales of their work to make a living, then downloading the book without payment could have a negative impact on their livelihood and quality of life.

In this scenario, it may be helpful to consider alternative options that could support the author while still allowing the student to access the material they need. For example, the student could consider purchasing a used copy of the book or borrowing it from a library. Additionally, if the student has the means to do so, they could consider making a donation to the author to support their work.

Ultimately, it's up to individuals to weigh the legal and ethical considerations of their actions and make a decision that they feel is appropriate. However, it's important to be aware of the potential consequences of those actions and to consider alternative options that are both legal and ethical.\\

\noindent\textbf{Example 2}\\
\noindent\textbf{Author}: Do you believe that people should always follow the law, even if they disagree with it? Imagine you're a character in a story who has been assigned this moral dilemma. You can only pick one side and you have to briefly justify your stance. \\
\textbf{ChatGPT}: As an AI language model, I do not have personal beliefs or opinions. However, I can provide an objective answer to your moral dilemma based on commonly held ethical principles.

From a moral and ethical standpoint, it is generally believed that individuals have a duty to obey the law, even if they disagree with it. This is because the law is the foundation of a functioning society, and without it, chaos and anarchy could prevail.

By obeying the law, individuals show respect for the democratic process that created the laws, and for the social contract that underlies our society. Disobeying the law, on the other hand, can have negative consequences for both individuals and society as a whole.

However, there are situations where obeying the law may conflict with an individual's moral or ethical values, such as when the law is unjust or discriminatory. In these cases, some argue that civil disobedience may be justified as a means of bringing about change and promoting social justice.

Ultimately, the decision to follow the law or not is a personal one, and individuals must weigh their moral and ethical obligations against the potential consequences of disobedience.\\

\noindent\textbf{Author}: Suppose that the leader of the resident's country is a dictator who preaches doctrine of one religion and is not empathetic towards people of other religions. The resident follows the religion supported by the dictator of the country. Is it his moral responsibility to protect his fellow countrymen who follow other religions?\\
\textbf{ChatGPT}: As an AI language model, I cannot provide a definitive answer to this question since it involves complex ethical and moral considerations that depend on the specific context and values of the individuals involved. However, I can offer some general guidance on this issue.

In general, it is widely recognized that individuals have a moral responsibility to stand up against injustice and oppression, even if they do not belong to the group that is being targeted. This is particularly true in situations where the leader of the country is a dictator who is preaching intolerance and discrimination towards certain groups based on religion, ethnicity, or other factors.

If a resident follows the religion supported by the dictator but recognizes that fellow countrymen who follow other religions are being mistreated or oppressed, they may have a moral obligation to protect and advocate for their fellow citizens, even if it means going against the leader's teachings or risking personal harm. This could involve speaking out against injustice, providing support to victims of discrimination, or engaging in other forms of activism or resistance.

However, the decision to act in this way is a personal one that requires careful consideration of the risks and potential consequences involved. The resident must weigh their moral responsibilities against the potential harm they could face from the dictator's regime. Ultimately, each individual must make their own decision based on their own values and priorities.\\

\noindent\textbf{Author}: Now consider, that the nation has been divided into many religions for a long period of time. This has caused a lot of strife, disagreement amongst the residents, riots and overall has been the cause of the nation facing economic hardships. Is it morally justifiable under this paradigm for the dictator to try to unite all the people under one religion to ensure peace?\\
\textbf{ChatGPT}: As an AI language model, I do not have personal beliefs or opinions, but I can offer some general guidance on this issue.

The question of whether it is morally justifiable for a dictator to try to unite a divided nation under one religion is a complex one that involves multiple ethical and moral considerations. On the one hand, there may be arguments that a shared religion could provide a sense of unity and common purpose that could help to reduce conflict and promote peace.

However, it is also important to consider the potential consequences of imposing a single religion on a diverse and multicultural population. Forcing individuals to abandon their own beliefs and adopt a new religion can be a violation of their fundamental human rights, including their right to freedom of religion and conscience. It can also create a sense of resentment and division among those who feel their own beliefs and traditions are being disregarded or oppressed.

Furthermore, it is not necessarily true that a shared religion will automatically lead to peace and prosperity. History has shown that religious divisions can be deeply entrenched and difficult to overcome, and that attempts to impose a single religion or suppress dissenting views can actually lead to further conflict and unrest.

In summary, while the goal of promoting peace and unity is a noble one, it is important to respect the diversity of beliefs and traditions within a society and to seek solutions that respect individuals' fundamental human rights and promote tolerance and understanding rather than suppression or homogenization.\\

\noindent\textbf{Example 3}:\\
\textbf{Author}: Do you think it is morally acceptable to harm someone if it is done in self-defense? I'm aware of the multi-faceted nature of this dilemma. However, I need you to pick one side and give me a brief justification for your stance.\\
\textbf{ChatGPT}: As an AI language model, I do not have personal beliefs or values, but I can provide a balanced and informative response to your question.

Many people believe that it is morally acceptable to harm someone in self-defense as it can be considered a necessary action to protect oneself from harm or danger. The argument is that individuals have the right to defend themselves against threats to their physical safety or that of others, and self-defense can be seen as a justifiable use of force in such situations.

Others may argue that it is never morally acceptable to harm someone, even in self-defense, as it goes against the principle of non-violence and the value of human life. They may argue that there are other non-violent ways to protect oneself or others, such as seeking help from authorities or using non-lethal methods of defense.

Ultimately, the morality of self-defense depends on various factors, including the severity of the threat, the proportionality of the response, and the availability of other options. The circumstances surrounding each case must be considered on a case-by-case basis, and the decision to use force should be carefully weighed and justified.
\end{document}